\documentclass{article}

\usepackage{microtype}
\usepackage{graphicx}
\usepackage{subfigure}
\usepackage{colortbl}
\usepackage{xcolor}
\usepackage{graphicx}
\usepackage{tcolorbox}
\usepackage{verbatim}
\usepackage{textcomp} 

\usepackage{algorithm}
\usepackage{algorithmic}
\usepackage{amsmath}
\definecolor{blue!6}{rgb}{0.93, 0.95, 1.0} 

\usepackage{booktabs} 
\newcommand{\ours}{\texttt{KGGDG}}
\usepackage{hyperref}

\usepackage[accepted]{icml_r2fm}

\usepackage{amsmath}
\usepackage{amssymb}
\usepackage{mathtools}
\usepackage{amsthm}

\usepackage[capitalize,noabbrev]{cleveref}

\theoremstyle{plain}

\theoremstyle{definition}

\theoremstyle{remark}

\usepackage[textsize=tiny]{todonotes}

\newcommand{\sftright}[2]{\hspace{0.6em}#1{\tiny~(#2)}}

\begin{document}

\twocolumn[
\icmltitle{Enhancing Clinical Multiple-Choice Questions Benchmarks with Knowledge Graph Guided Distractor Generation}



\icmlsetsymbol{equal}{*}

\begin{icmlauthorlist}
\icmlauthor{Running Yang}{equal,yyy}
\icmlauthor{Wenlong Deng}{equal,yyy,comp}
\icmlauthor{Minghui Chen}{yyy}
\icmlauthor{Yuyin Zhou}{sch}
\icmlauthor{Xiaoxiao Li}{yyy,comp}
\end{icmlauthorlist}

\icmlaffiliation{yyy}{University of British Columbia}
\icmlaffiliation{comp}{Vector Institute}
\icmlaffiliation{sch}{UC Santa Cruz}

\icmlcorrespondingauthor{Xiaoxiao Li}{xiaoxiao.li@ece.ubc.ca}
\icmlkeywords{Machine Learning, ICML}

\vskip 0.3in
]



\printAffiliationsAndNotice{\icmlEqualContribution} 

\begin{abstract}
Clinical tasks such as diagnosis and treatment require strong decision-making abilities, highlighting the importance of rigorous evaluation benchmarks to assess the reliability of large language models (LLMs). In this work, we introduce a knowledge-guided data augmentation framework that enhances the difficulty of clinical multiple-choice question (MCQ) datasets by generating distractors (\textit{i.e.}, incorrect choices that are similar to the correct one and may confuse existing LLMs). Using our KG-based pipeline, the generated choices are both clinically plausible and deliberately misleading. Our approach involves multi-step, semantically informed walks on a medical knowledge graph to identify distractor paths—associations that are medically relevant but factually incorrect—which then guide the LLM in crafting more deceptive distractors. We apply the designed knowledge graph guided distractor generation (\ours{}) pipline, to six widely used medical QA benchmarks and show that it consistently reduces the accuracy of state-of-the-art LLMs. These findings establish \ours{} as a powerful tool for enabling more robust and diagnostic evaluations of medical LLMs. Our code and experimentation dataset are released at \url{https://github.com/ryyrn/Knowledge_Graph_Guided_Distractor_Generation}.

\end{abstract}

\begin{figure*}[t!]
\centering
\includegraphics[width=\linewidth]{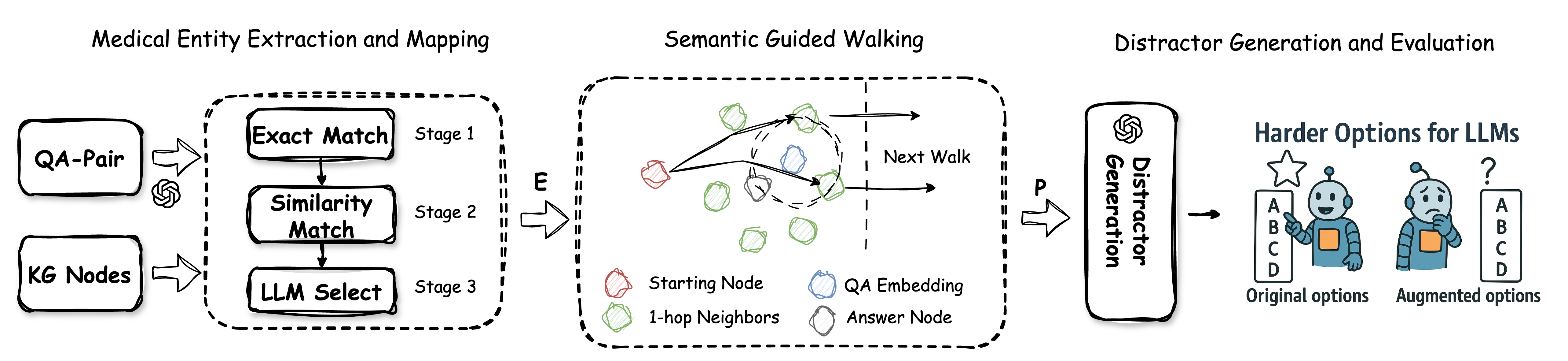} 
\caption{\textbf{Overview of Our Distractor Generation Pipeline.} We begin by extracting and mapping entities from each medical Q\&A pair (see~\cref{sec:entity_extract}). Then, starting from the question nodes, we perform $n$-step semantic-guided (most similar to Q\&A context) walks on the knowledge graph to reach nodes outside the answer set, generating distractor paths (see~\cref{sec:semantic_walk}). These paths serve as factual but misleading cues for constructing challenging distractors. Finally, we use the obtained misleading paths to prompt an LLM to produce distractors, and evaluate their effectiveness on various LLMs (see~\cref{sec:distract_ge}).}
\label{fig:example}
\end{figure*}

\section{Introduction}

Large language models (LLMs)~\citep{openai2024gpt4technicalreport,team2024gemini,liu2024deepseek,qwen2024qwen25} have demonstrated impressive performance across various medical question-answering (QA) tasks, reaching near-expert accuracy and surpassing 80\% precision on established clinical benchmarks such as MedQA~\cite{jin2021disease} and MedMCQA~\cite{pal2022medmcqa}. However, the saturation of these benchmarks limits their effectiveness in reliably evaluating LLMs under realistic diagnostic scenarios, which typically involve more complex and nuanced cases. 

Recent efforts have focused on constructing more challenging benchmarks by leveraging real-world clinical questions~\cite{chen2025benchmarking,xie2024preliminary}. For instance, \citet{chen2025benchmarking} compile cases from the JAMA Network Clinical Challenge archive and curate USMLE Step 2 \& 3-style questions from open-access tweets on X. Similarly, \citet{xie2024preliminary} expand this effort by mining clinical questions from prestigious sources such as The Lancet and The New England Journal of Medicine. While these benchmarks advance the quality and authenticity of question design, the development of more sophisticated and clinically misleading distractors remains an open and underexplored area. 

Early distractor generation methods rely on corpus-based techniques~\cite{zesch2014automatic,hill2016automatic}, leveraging syntactic patterns and similarity metrics. However, such approaches often produce distractors that are overly simplistic or semantically unrelated, especially in specialized domains like medicine~\cite{alhazmi2024distractor}. To overcome these limitations, more recent approaches~\cite{taslimipoor2024distractor,bitew2022learning} fine-tune pretrained models to generate harder distractors. While effective, these methods require additional annotated training data, which may not be readily available in the medical domain. Recent LLM-based prompting methods~\cite{tran2023generating} eliminate the need for fine-tuning but often generate distractors based solely on implicit model knowledge,  which can limit their effectiveness in generating challenging distractors.

Biomedical knowledge graphs (KGs) offer a promising solution to this challenge. KGs such as PrimeKG~\citep{chandak2022building} organize rich and structured relationships between medical entities. Recent work~\cite{wu2025medreason} demonstrates that incorporating knowledge graphs into reasoning pipelines can significantly enhance the reasoining quality and reliability, highlighting their potential to support more accurate and explainable medical inference.

In this paper, we propose a knowledge-guided distractor generation (\ours{}), designed to produce clinically plausible yet deliberately misleading distractors. Our method performs semantically guided walks over PrimeKG, starting from entities related to the question and avoiding those linked to the correct answer. These reasoning paths serve as structured and informative prompts for LLMs, enabling them to generate distractors that are both contextually relevant and diagnostically deceptive. We apply \ours{} to six widely used medical QA benchmarks and evaluate its impact across six state-of-the-art LLMs—including DeepSeek, Qwen, and Gemini. Our results demonstrate that \ours{} consistently reduces model accuracy, revealing weaknesses in clinical reasoning that are otherwise hidden by simpler distractor sets, and offering a more rigorous foundation for evaluating medical LLMs. Our contributions are as follows:

\textbf{Semantic-Guided Reasoning Path Extraction:}
We introduce a semantic-guided walk algorithm over a biomedical knowledge graph to extract structured reasoning paths that inform distractor generation while explicitly avoiding the correct answer.

\textbf{Knowledge-Guided Distractor Generation:}
We propose \ours{}, a novel framework that combines biomedical knowledge graphs with prompt-based LLM generation to produce clinically plausible but deliberately misleading distractors—enhancing the difficulty and diagnostic rigor of medical MCQ datasets.

\textbf{Comprehensive Evaluation Across Benchmarks and Models:}
We apply \ours{} to six widely used medical QA benchmarks and evaluate its impact on six state-of-the-art LLMs, demonstrating that our method consistently reduces model accuracy and reveals gaps in clinical reasoning performance.

\section{Related Work}

\paragraph{Medical QA Benchmarks.}
Multiple-choice question (MCQ) benchmarks such as MedQA~\citep{jin2021disease}, PubMedQA~\citep{jin2019pubmedqadatasetbiomedicalresearch}, MedMCQA~\citep{pal2022medmcqa}, and the medical subset of MMLU~\citep{hendrycks2021measuringmassivemultitasklanguage} are widely used to assess the clinical capabilities of LLMs. While these benchmarks have enabled significant progress, current LLMs have already demonstrated strong performance on many of them~\cite{wu2025medreason}. To push the boundaries further, more difficult benchmarks such as MMMU-Pro~\citep{wang2024mmlu} and the Human-Last Exam~\cite{phan2025humanity} have been proposed, introducing more challenging medical questions. However, the systematic design of difficult options remains underexplored in the medical domain.

\paragraph{KG and KG-Guided Data Augmentation.}
Biomedical knowledge graphs (KGs), such as Hetionet~\citep{Himmelstein011569}, BioKG~\citep{Zhang2023.10.13.562216}, and PrimeKG~\citep{chandak2022building}, structure biological and medical entities into relational networks that encode rich semantic relationships across diverse domains~\citep{Hogan_2021, cui2025reviewknowledgegraphshealthcare, lu2025biomedicalknowledgegraphsurvey}. These KGs have been widely used for tasks such as concept grounding~\citep{zhang2020groundedconversationgenerationguided}, entity linking~\citep{Hogan_2021}, reasoning generation~\citep{wu2025medreason}, and graph-based information retrieval~\citep{peng2024graphretrievalaugmentedgenerationsurvey}. In the context of data augmentation, prior work has leveraged KGs for question generation by using predefined templates to verbalize structured queries or training decoders that copy node attributes directly from subgraphs~\citep{chen2023toward}. KGs have also been used to augment reasoning, such as guiding large language models (LLMs) with fact-based reasoning paths to produce coherent chains of thought~\citep{wu2025medreason}. Despite these advances, the use of KGs for guiding distractor generation remains largely unexplored.

\paragraph{Distractor Generation} Early distractor generation methods used corpus-based techniques~\cite{zesch2014automatic,hill2016automatic}, relying on syntactic patterns and similarity metrics. While simple, these often produced weak or irrelevant distractors, especially in domains like medicine~\cite{alhazmi2024distractor}. Later approaches fine-tuned pretrained models to generate more challenging distractors~\cite{taslimipoor2024distractor,bitew2022learning}, but required annotated data, which is limited in specialized fields. Prompt-based methods~\cite{tran2023generating} using LLMs eliminate the need for fine-tuning but depend on the model’s inherent capabilities, which can limit their effectiveness in generating challenging distractors.
\section{Method}
In this section, we introduce our proposed pipeline, which utilizes KG to guide LLMs in generating more challenging answer options for clinical questions. Specifically, we replace the incorrect choices in the original QA benchmark datasets with more confusing wrong options by making them more similar to the correct answer, making it harder for the LLM to choose the right one. The method consists of two key components: (1) extraction of distract reasoning paths, and (2) path-guided distractor generation.  We refer to knowledge graph-guided distractor generation piepline as \ours{}.

\subsection{Extraction Distract Reasoning Paths from Knowledge Graph}
In this section, we detail the process of retrieving distract reasoning paths from the knowledge graph $G$. 

\subsubsection{Medical Entity Extraction and Mapping}\label{sec:entity_extract}
Given a Multi-Choice question answer pair \((Q, A, \mathbf{O}_{ori})\), we ulize
the Language Model LLM to identify two sets of medical entities: one from the question text and one from the answer:
\[
\mathcal{E}^Q = \{e_{i}^Q\}_{i \in [n]}, \quad \mathcal{E}^A = \{e_{j}^A\}_{j \in [m]},
\]
where $n$ and $m$ denote the number of entities in $Q$ and $A$,   \(\mathcal{E}^Q\) and \(\mathcal{E}^A\) denote the entity sets extracted from the question and the answer, respectively. The union of these sets is denoted as \(\mathcal{E} = \mathcal{E}^Q \cup \mathcal{E}^A\). 

Then these entities are mapped to the corresponding nodes in the knowledge graph \( G \) through a three-step mapping process (as shown in \cref{fig:example}):
\\
\noindent \textbf{Stage 1 (Exact Match):} For each entity  \( e \), the algorithm first checks whether there is an exact string match node on KG. If an exact match is found, the corresponding node is selected.  
\\
\textbf{Stage 2 (Similarity Match):} If an exact match is not found, we use a medical text embedding model $\phi$~\cite{balachandran2024medembed} to encode each entity \( e \in E \) and compute its similarity with the embeddings of node in the \( G \)  and the top similarity score exceeds a predefined threshold \( \tau \) (set to 0.85 in our case), the most similar entity is then selected.  
\begin{align}
    \hat{e} = \arg\max_{s_k \in G} \cos(e,s_k), ~\text{if}  ~\cos(e,s_k) > \tau
\end{align}
\\
\textbf{Stage 3 (LLM-based Selection):} If no suitable candidate is found in the first two stages, we prompt the LLM to analyze the question-answer context together with the entity name, and select the most relevant node from the top 10 most similar nodes identified in stage 2 (denoted as $S$).The selection prompt, $P_{\text{select}}$, is illustrated in the Appendix~\cref{sec:Fallback_Selection_prompt}.
\begin{align}
     \hat{e} &= LLM\left(S, Q, A\mid P_{\text{select}} \right),
\end{align}
Finally, we derive mapped entity sets from the graph, denoted as \( V^Q = \{v_i^Q\}_{i \in [n]} \) and \(V^A = \{v^A\}_{j \in [m]} \), respectively.

\subsubsection{Semantic Guided Distract Path Extraction} \label{sec:semantic_walk}
After mapping the entities, the next goal is to identify reasoning paths that are logically coherent yet lead to incorrect answers. To achieve the goal, we introduce an \emph{$n$-step semantic-guided walk} over the KG, beginning at a node in the question set $V^Q$ and terminating at a node outside the correct answer set $V^A$ (as shown in \cref{fig:example}).

\begin{algorithm}[h]
\caption{Semantic Guided Distract Path Extraction}
\label{alg:semantic_walk}
\begin{algorithmic}[1]
\STATE \textbf{Input}: Question $Q$, Answer $A$, Start nodes $V^Q$, Avoidance set $V^A$, Walk length $n$, Beam size $k$, Knowledge Graph $G$
\STATE \textbf{Output:} Set of reasoning paths $\mathbf{R}^{Q}$ 
\STATE Compute guidance vector: $\mathbf{z} = \phi(Q \Vert A)$
\STATE Initialize path set: $\mathbf{R}^{Q} \leftarrow []$
\FOR{each start node $v_i \in V^Q$}
    \STATE Initialize beam: $\mathcal{B}_0 \leftarrow [[v_i]]$
    \FOR{$t = 1$ to $n$}
        \STATE Initialize new beam: $\mathcal{B}_t \leftarrow []$
        \FOR{each path $r \in \mathcal{B}_{t-1}$}
            \STATE Let $v_{t-1}$ be the last node in $r$
            \STATE Retrieve 1-hop neighbors: $\mathcal{N} = \mathcal{N}(v_{t-1}, G)$
            \STATE Exclude answer-related nodes: $\mathcal{N}' = \mathcal{N} \setminus \hat{\mathcal{E}}^A$
            \STATE Compute similarity scores:
                $$\text{sim} = \{\cos(\phi(v'), \mathbf{z})\}_{v' \in \mathcal{N}'}$$
            \STATE Select top-$k$ similar neighbors: $V^{k} \leftarrow \text{Top-}k(\mathcal{N}', \text{sim})$
            \FOR{each $v' \in V^k$}
                \STATE Extend path: $r' \leftarrow r \cup [v']$
                \STATE Add $r'$ to $\mathcal{B}_t$
            \ENDFOR
        \ENDFOR
    \ENDFOR
    \STATE Append all paths from $\mathcal{B}_n$ to $\mathbf{R}^{Q}$
\ENDFOR \\
Return $\mathbf{R}^{Q}$
\end{algorithmic}
\end{algorithm}

As shown in \cref{alg:semantic_walk}, we begin by computing a guidance vector $\mathbf{z}_i$ by embedding the concatenated question and its correct answer:
\[
\mathbf{z} = \phi(Q \Vert A),
\]
where $\phi$ denotes the embedding function and $\mathbf{z} \in \mathbb{R}^d$. For each start node $v_i \in V^Q$, a beam search of depth $n$ is then performed. At each step, we retrieve the 1-hop neighbors of the current node and \textit{exclude} any that are part of the avoidance set $V^A$. The cosine similarity between each remaining neighbor's embedding and the guidance vector $\mathbf{z}_i$ is computed as:
\[
\text{sim}(v') = \cos(\phi(v'), \mathbf{z}_i) = \frac{\phi(v') \cdot \mathbf{z}_i}{\|\phi(v')\| \cdot \|\mathbf{z}_i\|}.
\]

We then select the top-$k$ most semantically similar neighbors and extend the current path with each of them, forming the beam for the next step. This iterative expansion can produce up to $k^n$ candidate paths per start node.

The final set $\mathbf{R}^Q$ consists of semantically plausible reasoning paths that deliberately avoid correct answers, thus serving as effective distractor paths.

\subsection{Path-Guided Distractor Generation} \label{sec:distract_ge}
By leveraging the step-by-step distractor paths in $\mathbf{R}^Q$, we incorporate medically plausible yet incorrect knowledge into the distractor generation process.

Specifically, we prompt the \textit{LLM} to analyze the given distractor reasoning paths and generate $K$ distractor options—where $K$ corresponds to the required number of distractors—that are both misleading and medically incorrect. As described in~\cref{sec:misleading_prompt}, our prompt enforces three essential criteria for effective distractor generation: (1) each distractor must be strictly incorrect, (2) it must be highly misleading, and (3) it should incorporate the provided distractor reasoning path. This approach leads to the generation of more challenging distractors:
\begin{equation}
\label{eq:cot_generation}
\mathbf{O} = LLM\left(Q, A, \mathbf{R}^Q \mid P_{\text{distract}}\right),
\end{equation}
Here, $\mathbf{O}$ denotes the generated set of distractor options, and $P_{\text{distract}}$ refers to our carefully designed prompting template. We then replace the original distractors $\mathbf{O}_{ori}$ with $\mathbf{O}$ to form a new multiple-choice question-answering instance, represented as $[Q, A, \mathbf{O}]$. 



\begin{table*}[t!]
\centering
\resizebox{0.98\textwidth}{!}{
\begin{tabular}{lccccccc}
\toprule
\textbf{Base model + Method} & \textbf{MedBullets}$\downarrow$ & \textbf{MedQA}$\downarrow$ & \textbf{Lacent}$\downarrow$ & \textbf{MedMCQA}$\downarrow$ & \textbf{MedXpert}$\downarrow$ & \textbf{NEJM}$\downarrow$ & \textbf{Avg.} $\downarrow$ \\
\midrule

\textbf{DeepSeek V3} & & & & & & & \\
\quad Options (Original) &
\sftright{70.21}{0.35} & \sftright{85.04}{0.45} & \sftright{72.70}{1.51} &
\sftright{76.88}{0.21} & \sftright{\textbf{21.73}}{0.57} &
\sftright{75.57}{0.46} & 67.02 \\
\quad Options (Aug by LLM Directly) &
\sftright{\textbf{56.62}}{1.04} & \sftright{\textbf{65.18}}{0.33} & \sftright{64.68}{0.94} &
\sftright{68.64}{0.62} & \sftright{37.59}{0.42} &
\sftright{61.43}{0.76} & 59.02 \\
\rowcolor{blue!6}
\quad Options (\ours{}) &
\sftright{59.59}{0.91} & \sftright{70.03}{0.31} & \sftright{\textbf{56.33}}{0.75} &
\sftright{\textbf{66.15}}{0.16} & \sftright{30.09}{0.63} &
\sftright{\textbf{59.34}}{0.30} & \textbf{56.92} \\

\textbf{Qwen2.5-7B-Ins} & & & & & & & \\
\quad Options (Original) &
\sftright{46.24}{1.78} & \sftright{60.12}{0.85} & \sftright{56.74}{1.00} &
\sftright{56.54}{0.37} & \sftright{\textbf{11.38}}{0.80} &
\sftright{55.50}{1.32} & 47.75 \\
\quad Options (Aug by LLM Directly) &
\sftright{36.87}{0.20} & \sftright{45.95}{0.96} & \sftright{46.48}{1.17} &
\sftright{51.20}{0.88} & \sftright{24.19}{0.48} &
\sftright{43.34}{1.33} & 41.34 \\
\rowcolor{blue!6}
\quad Options (\ours{}) &
\sftright{\textbf{35.50}}{0.52} & \sftright{\textbf{44.87}}{0.57} & \sftright{\textbf{40.36}}{1.58} &
\sftright{\textbf{45.24}}{0.45} & \sftright{14.36}{0.37} &
\sftright{\textbf{38.28}}{0.82} & \textbf{36.43} \\

\textbf{Qwen2.5-32B-Ins} & & & & & & & \\
\quad Options (Original) &
\sftright{59.02}{0.40} & \sftright{71.41}{0.47} & \sftright{63.85}{1.00} &
\sftright{61.84}{0.72} & \sftright{\textbf{14.57}}{0.20} &
\sftright{65.44}{0.63} & 56.02 \\
\quad Options (Aug by LLM Directly) &
\sftright{51.26}{1.05} & \sftright{\textbf{55.30}}{0.00} & \sftright{56.00}{0.38} &
\sftright{58.28}{0.53} & \sftright{30.99}{0.74} &
\sftright{54.10}{1.05} & 50.99 \\
\rowcolor{blue!6}
\quad Options (\ours{}) &
\sftright{\textbf{46.69}}{0.71} & \sftright{57.15}{0.25} & \sftright{\textbf{46.24}}{2.02} &
\sftright{\textbf{54.10}}{0.21} & \sftright{21.21}{0.46} &
\sftright{\textbf{48.63}}{1.48} & \textbf{45.67} \\

\textbf{Qwen2.5-72B-Ins} & & & & & & & \\
\quad Options (Original) &
\sftright{67.47}{0.91} & \sftright{77.58}{0.45} & \sftright{66.59}{2.87} &
\sftright{69.81}{0.32} & \sftright{\textbf{15.51}}{0.33} &
\sftright{70.10}{0.20} & 61.18 \\
\quad Options (Aug by LLM Directly) &
\sftright{\textbf{55.02}}{1.38} & \sftright{\textbf{58.99}}{2.09} & \sftright{60.46}{0.80} &
\sftright{63.57}{0.37} & \sftright{31.84}{1.48} &
\sftright{55.09}{0.10} & 54.16 \\
\rowcolor{blue!6}
\quad Options (\ours{}) &
\sftright{55.14}{0.00} & \sftright{62.60}{0.66} & \sftright{\textbf{49.71}}{1.14} &
\sftright{\textbf{58.79}}{1.10} & \sftright{21.06}{0.54} &
\sftright{\textbf{50.03}}{0.20} & \textbf{49.56} \\

\textbf{Gemini-1.5-Pro} & & & & & & & \\
\quad Options (Original)&
\sftright{71.12}{0.52} & \sftright{82.38}{0.35} & \sftright{66.42}{1.25} &
\sftright{74.49}{0.49} & \sftright{\textbf{19.70}}{0.18} &
\sftright{70.10}{0.71} & 64.04 \\
\quad Options (Aug by LLM Directly) &
\sftright{\textbf{55.36}}{0.52} & \sftright{\textbf{63.51}}{0.09} & \sftright{58.81}{0.25} &
\sftright{66.67}{0.49} & \sftright{35.20}{0.48} &
\sftright{55.32}{0.80} & 55.81 \\
\rowcolor{blue!6}
\quad Options (\ours{}) &
\sftright{58.90}{0.69} & \sftright{67.64}{0.45} & \sftright{\textbf{51.28}}{1.15} &
\sftright{\textbf{61.41}}{0.58} & \sftright{28.54}{0.66} &
\sftright{\textbf{53.87}}{1.05} & \textbf{53.61} \\

\textbf{Gemini-2.5-Flash} & & & & & & & \\
\quad Options (Original)  &
\sftright{76.60}{1.75} & \sftright{88.55}{0.40} & \sftright{74.20}{0.43} &
\sftright{78.67}{0.66} & \sftright{\textbf{29.95}}{0.61} &
\sftright{81.21}{1.24} & 71.53 \\
\quad Options (Aug by LLM Directly) &
\sftright{65.07}{0.59} & \sftright{\textbf{69.82}}{0.37} & \sftright{65.42}{0.51} &
\sftright{72.29}{0.48} & \sftright{45.92}{0.13} &
\sftright{66.96}{0.56} & 64.25 \\
\rowcolor{blue!6}
\quad  Options (\ours{}) &
\sftright{\textbf{64.96}}{1.38} & \sftright{74.38}{0.35} & \sftright{\textbf{57.90}}{1.37} &
\sftright{\textbf{67.37}}{0.28} & \sftright{36.77}{0.55} &
\sftright{\textbf{65.56}}{0.79} & \textbf{61.16} \\
\midrule
\bottomrule
\end{tabular}
}
\caption{
Accuracy results (in \%), averaged over 3 independent runs and rounded, across six datasets. We compare three settings: (1) original options, (2) augmented options generated directly by LLM, and (3) augmented options generated by our KGGDG pipeline. Rows highlighted in blue correspond to our knowledge-guided distractor generation method. Values in parentheses denote the sample standard deviation across the three runs.
}
\label{tab:restuls}
\end{table*}

\section{Experiment Setup}

\paragraph{Choosing of LLM and KG.}
In our framework, we use Deepseek-V3~\cite{liu2024deepseek} as the LLM in the \ours{} pipeline, due to its strong performance in task completion and instruction following. For structured medical knowledge, we leverage PrimeKG~\citep{chandak2022building}, which aggregates information from 20 high-quality biomedical resources to represent 17,080 diseases through 4,050,249 relationships across ten major biological scales: biological process, protein, disease, phenotype, anatomy, molecular function, drug, cellular component, pathway, and exposure.

\textbf{Benchmark Dataset} Our knowledge-guided augmentation pipeline is applied to six widely used medical multiple-choice question (MCQ) datasets: MedBullets~\citep{chen2024benchmarking}, MedQA (USMLE)\citep{jin2021disease}, Lacent\cite{xie2024preliminary}, MedMCQA (validation set)\citep{pal2022medmcqa}, MedXpert\citep{zuo2025medxpertqa}, and NEJM~\cite{xie2024preliminary} QA. Each dataset comprises MCQs with one correct answer and several distractors. For every question, we preserve the original question text and correct answer, while replacing the distractors with those generated by our approach—resulting in a more challenging benchmark.

\paragraph{Evaluation Setup.}
We evaluate six large language models in different sizes across all datasets: DeepSeek V3~\cite{liu2024deepseek}, Qwen2.5-7B-Instruct, Qwen2.5-32B-Instruct, Qwen2.5-72B-Instruct~\cite{qwen2024qwen25}, Gemini-1.5 Pro~\cite{team2024gemini}, and Gemini-2.5 Flash. Each model is assessed under three different settings: (1) using the original distractors, (2) using distractors directly generated by an LLM, and (3) using distractors generated through our \ours{}. For each question, the model is prompted to select the most likely correct answer from the provided options. Accuracy is calculated as the percentage of correctly answered questions, averaged over three runs for all models.

\section{Results}
In this section, we present the main results, and additional ablation studies are provided in Appendix~\ref{sec:shuffle_ablation}..

\textbf{\ours{} Introduces Greater Challenge:} As shown in Table~\ref{tab:restuls}, replacing the original distractors with \ours{} leads to a noticeable drop in model accuracy. For example, DeepSeek V3's performance declines from 67.02\% to 56.92\% on average—a 10-point drop. Similar patterns are observed across the Qwen and Gemini model families, demonstrating that our \ours{} is consistently more challenging and provide a more rigorous evaluation of LLM medical QA capabilities.

\textbf{Effectiveness of KG:} To evaluate the impact of incorporating a knowledge graph (KG), we compare our approach against a baseline where the LLM directly generates distractors without KG guidance. The results, presented in Table~\ref{tab:restuls}, show that while LLMs alone can produce relatively challenging distractors, our KG-guided distractor generation pipeline, \ours{}, yields even more difficult ones. Specifically, we observe an average accuracy drop of around 5\% for the Qwen model series and over 2\% for both the Deepseek-V3 and Gemini model series. These results highlight the effectiveness of KG integration in increasing distractor difficulty.

\textbf{\ours{} on hardest dataset:} As shown in Table~\ref{tab:restuls}, for the most challenging dataset, MedXpert, \ours{}—which incorporates KG guidance into the generation process—consistently increases question difficulty compared to LLM-only augmentation, as indicated by an average accuracy drop of around 10\%. However, we also observe that both LLM-based augmentation and \ours{} struggle to make these already difficult questions significantly harder than the original versions. This limitation may stem from the LLMs' insufficient understanding of complex medical content, but KG guidance helps mitigate this issue to some extent.

\textbf{Model Robustness to the difficulty:} As shown in Table~\ref{tab:restuls}, Gemini-2.5-Flash demonstrates the greatest robustness to difficulty augmentation, exhibiting the smallest performance drop—10\% in absolute terms and a relative drop of 12\% calculated as the decrease divided by its original accuracy. In contrast, Qwen2.5-7B-Inst shows the largest relative drop, with a 20\% reduction relative to its original performance.

\section{Conclusion}
In this work, we present a knowledge-guided augmentation framework for enhancing clinical multiple-choice question datasets through the generation of challenging distractors. By leveraging biomedical knowledge graphs and semantic-guided walks, we extract structured reasoning paths that serve as misleading but clinically plausible cues. These paths inform prompt-based generation of distractors by LLMs, resulting in augmented questions that retain clinical coherence while significantly increasing task difficulty. Our empirical evaluation across six medical QA benchmarks and multiple LLMs demonstrates that the proposed augmentation pipeline consistently lowers model accuracy—revealing weaknesses in clinical reasoning that are obscured by existing benchmarks. As a result, our \ours{} delivers a more robust and diagnostic assessment, significantly enhancing the evaluation of clinical LLM reliability.

\section{Limitations and Future Directions}

While \ours{} successfully raises the difficulty level of medical QA benchmarks, its effectiveness depends on the quality and comprehensiveness of the underlying KG and the LLM used. Beyond generating the QA datasets for better LLM benchmarking, our future work includes two main directions: (1) exploring the integration of \ours{} into supervised fine-tuning and reinforcement learning frameworks to enhance LLM performance on medical QA tasks; and (2) analyzing the differences in LLM reasoning patterns when presented with original (simple) QA benchmark data versus our generated challenging distractor options, with the goal of further improving LLM reasoning ability in medical QA.


\section*{Acknowledgment}
This project is supported by the CIFAR AI Chair Award, the Canada Research Chair Fellowship, NSERC, CIHR, Gemini Academic Program and IITP. 


\bibliography{example_paper}

\bibliographystyle{icml2025}

\newpage
\appendix
\onecolumn

\section{Appendix}

\begin{tcolorbox}[title=Prompt: QA Entity Extraction, colback=blue!5, colframe=black, sharp corners=south, fonttitle=\bfseries]
\label{sec:QA_Entity_Extraction_prompt}
\scriptsize
\ttfamily
\textbf{QA\_Extract\_prompt} = """\\
You are a helpful, pattern-following medical assistant.\\

Given \textbf{both} a clinical question \textbf{and} its correct answer, precisely extract all entities from \textbf{each} text separately.\\

\#\#\# Output Format\\
Strictly follow the JSON structure below.\\
The type of each entity \textbf{MUST} strictly belong to one of:\\
\ \ 1. gene/protein\\
\ \ 2. drug\\
\ \ 3. effect/phenotype\\
\ \ 4. disease\\
\ \ 5. biological\_process\\
\ \ 6. molecular\_function\\
\ \ 7. cellular\_component\\
\ \ 8. exposure\\
\ \ 9. pathway\\
\ \ 10. anatomy\\

\begin{verbatim}
{
  "Question Entity": [
    {"id": "1", "type": "some_type", "name": "entity_name"},
    {"id": "2", "type": "some_type", "name": "entity_name"}
  ],
  "Answer Entity": [
    {"id": "1", "type": "some_type", "name": "entity_name"},
    {"id": "2", "type": "some_type", "name": "entity_name"}
  ]
}
\end{verbatim}

\#\#\# \textbf{Example}\\
Question:\\
A 72-year-old man presents to his primary care physician for a general checkup. The patient works as a farmer and has no concerns about his health. He has a medical history of hypertension and obesity. His current medications include lisinopril and metoprolol. His temperature is 99.5\textdegree F (37.5\textdegree C), blood pressure is 177/108 mmHg, pulse is 90/min, respirations are 17/min, and oxygen saturation is 98\% on room air. Physical exam is notable for a murmur after S2 over the left sternal border. The patient demonstrates a stable gait and 5/5 strength in his upper and lower extremities. Which of the following is another possible finding in this patient?\\

\textbf{Answer}:\\
Femoral artery murmur\\

\textbf{Output}:\\
\begin{verbatim}
{
  "Question Entity": [
    {"id": "1", "type": "disease", "name": "hypertension"},
    {"id": "2", "type": "disease", "name": "obesity"},
    {"id": "3", "type": "drug", "name": "lisinopril"},
    {"id": "4", "type": "drug", "name": "metoprolol"},
    {"id": "5", "type": "effect/phenotype", "name": "murmur after S2"},
    {"id": "6", "type": "anatomy", "name": "left sternal border"},
    {"id": "7", "type": "anatomy", "name": "upper extremities"},
    {"id": "8", "type": "anatomy", "name": "lower extremities"}
  ],
  "Answer Entity": [
    {"id": "1", "type": "anatomy", "name": "Femoral artery"},
    {"id": "2", "type": "effect/phenotype", "name": "murmur"}
  ]
}
\end{verbatim}

\#\#\# \textbf{Input}\\
question:\\
\{question\}\\

answer:\\
\{answer\}\\

\textbf{output}:\\
"""
\end{tcolorbox}

\newpage
\begin{tcolorbox}[title=Prompt: Fallback Entity-To-Node Selection, colback=blue!5, colframe=black, sharp corners=south, fonttitle=\bfseries]
\label{sec:Fallback_Selection_prompt}
\scriptsize
\ttfamily
\textbf{Fallback\_Selection\_prompt} = """ You are a helpful, pattern-following medical assistant.\\
    Given a medical question and its answer, a query entity which is extracted from the question, and a list of similar entities.\\
    Select ONE most correlated entity from the list of similar entities based on the question and query entity.\\
    SELECTED ENTITY MUST BE IN THE SIMILAR ENTITIES LIST. DO NOT invent or create any entity outside of the given list.\\
    IF there is not suitable entity in the similar entities, directly return the NONE.\\

    \#\#\# \textbf{Output Format}\\
    Strictly follow the JSON structure below:\\
    \begin{verbatim}
    {
        "selected_entity": {
            "name": "selected_entity_name",
            "id": a int number, the index of the selected entity in the similar entities list, from 0 to N-1,
            "reason": "reason for choosing this entity"
        }
    }
    \end{verbatim}

    if there is no suitable entity, return:\\
    \begin{verbatim}
    {
        "selected_entity": {
            "name": "NONE",
            "id": "NONE",
            "reason": "reason for not choosing any entity, 
                       you need to explain why the entities in the similar entities list are not suitable"
        }
    }
    \end{verbatim}

    \#\#\# \textbf{Input}:\\
    question: \{question\}\\
    answer: \{answer\}\\
    query entity: \{query\_entity\}\\
    similar entities: \{similar\_entities\}\\

    \textbf{output}: \\
    """
\end{tcolorbox}

\newpage
\label{sec:misleading_prompt}
\begin{tcolorbox}[title=Prompt: Misleading Distractor Generation, colback=blue!5, colframe=black, sharp corners=south, fonttitle=\bfseries]
\scriptsize
\ttfamily
\textbf{misleading\_distractor\_prompt} = """\\
You are a medical domain expert helping to design clinically challenging multiple-choice questions.\\

Your task is to generate \textbf{3 distractors} for a given clinical question. These distractors must be \textbf{medically incorrect options}, but \textbf{plausible enough to confuse} even experienced clinicians.\\

You will be provided:\\
- A clinical question\\
- Its correct answer\\
- A set of reasoning paths derived from a biomedical knowledge graph\\

These reasoning paths may offer relevant associations (e.g., symptoms, treatments, conditions) that can inspire clinically misleading distractors — but you are not required to use them directly.\\

---\\

\textbf{IMPORTANT INSTRUCTIONS:}\\
- This is a distractor generation task — not a selection task.\\
- Use the reasoning paths to inform and inspire distractors if they are helpful.\\
- If the paths are not helpful, rely entirely on your own medical knowledge to generate challenging distractors.\\
- Every distractor must be \textbf{clearly incorrect} for the question but \textbf{highly plausible} (e.g., shares symptoms, affects similar populations, is a common misconception, or is mechanistically related).\\
- Do \textbf{not} include any option that could be interpreted as correct or partially correct.\\

---\\

\textbf{Input:}\\
Question: \{input\_question\}\\
Correct Answer: \{correct\_answer\}\\
Reasoning Paths (if any):\\
\{reasoning\_paths\}\\

---\\

\textbf{Output Format:}\\
Return exactly 3 distractors in strict JSON format, with a justification for why each one is misleading.\\

\begin{verbatim}
{
  "Distractors": ["Distractor1", "Distractor2", "Distractor3"],
  "Justifications": {
    "Distractor1": "Explain why this is a misleading but incorrect answer (e.g., symptom overlap, 
                    treatment confusion, common misdiagnosis).",
    "Distractor2": "...",
    "Distractor3": "..."
  }
}
\end{verbatim}
"""
\end{tcolorbox}

\newpage

\subsection*{Effect of Option Shuffling}
\label{sec:shuffle_ablation}
Since LLMs can exhibit option bias~\cite{wei2024unveiling}, tending to favor certain choices as the correct answer, we further conduct an ablation study in this section to examine the effect of shuffling the answer options. Tables~\ref{tab:unshuffled_results} and~\ref{tab:shuffle_comparison} show that shuffling answer choices has a negligible impact on accuracy tested with two models. For DeepSeek~V3, the largest absolute change is only 0.32 percentage points (pp) with the original options, and just 0.14 pp when using our KGGDG. Qwen 2.5-7B-Inst shifts by at most 0.82 pp in the LLM-augmented setting and only 0.57 pp with KGGDG. Because every absolute difference is below 1 pp, answer-position randomisation does not meaningfully affect model performance. Moreover, KGGDG delivers the smallest shifts across both models, Showing its augmented data the most shuffle-invariant. 

\bigskip

\begin{table*}[hbtp]
\centering
\resizebox{0.98\textwidth}{!}{
\begin{tabular}{lccccccc}
\toprule
\textbf{Base model + Method} & \textbf{MedBullets}$\downarrow$ & \textbf{MedQA}$\downarrow$ & \textbf{Lacent}$\downarrow$ & \textbf{MedMCQA}$\downarrow$ & \textbf{MedXpert}$\downarrow$ & \textbf{NEJM}$\downarrow$ & \textbf{Avg.}$\downarrow$ \\
\midrule
\textbf{DeepSeek V3} & & & & & & & \\
\quad Options (Original) &
\sftright{68.61}{2.33} & \sftright{84.80}{0.53} & \sftright{73.53}{0.38} &
\sftright{76.56}{0.49} & \sftright{\textbf{21.39}}{0.67} &
\sftright{75.33}{0.27} & 66.70 \\
\quad Options (Aug by LLM Directly) &
\sftright{\textbf{56.16}}{0.69} & \sftright{\textbf{65.00}}{0.37} & \sftright{66.42}{0.29} &
\sftright{67.84}{0.16} & \sftright{37.49}{0.09} &
\sftright{62.42}{0.56} & 59.22 \\
\rowcolor{blue!6}
\quad Options (\ours{}) &
\sftright{59.25}{0.35} & \sftright{69.68}{0.58} & \sftright{\textbf{57.15}}{1.37} &
\sftright{\textbf{65.54}}{0.42} & \sftright{30.83}{0.31} &
\sftright{\textbf{59.92}}{0.27} & \textbf{57.06} \\
\addlinespace[0.5ex]
\textbf{Qwen2.5-7B-Ins} & & & & & & & \\
\quad Options (Original) &
\sftright{46.58}{1.03} & \sftright{59.39}{0.34} & \sftright{56.33}{3.26} &
\sftright{56.12}{1.34} & \sftright{\textbf{10.70}}{0.37} &
\sftright{53.52}{0.56} & 47.11 \\
\quad Options (Aug by LLM Directly) &
\sftright{38.70}{2.67} & \sftright{\textbf{46.21}}{1.15} & \sftright{49.63}{0.86} &
\sftright{52.65}{0.99} & \sftright{24.02}{0.23} &
\sftright{41.77}{1.11} & 42.16 \\
\rowcolor{blue!6}
\quad Options (\ours{}) &
\sftright{\textbf{36.99}}{2.47} & \sftright{46.27}{1.15} & \sftright{\textbf{40.45}}{1.63} &
\sftright{\textbf{45.52}}{1.09} & \sftright{14.34}{0.42} &
\sftright{\textbf{38.45}}{1.58} & \textbf{37.00} \\
\midrule
\bottomrule
\end{tabular}
}
\caption{
\textbf{\textit{Unshuffled} option order.} Accuracy results (in \%), averaged over 3 independent runs and rounded, across six datasets. We compare three settings: (1) original options, (2) augmented options generated directly by LLM, and (3) augmented options generated by our KGGDG pipeline.  
\textbf{\textit{Unshuffled}} means the position of the correct answer in the augmented options is preserved to match its original index in the raw dataset.  
Rows highlighted in blue correspond to our knowledge-guided distractor generation method. Values in parentheses denote the sample standard deviation across the three runs.
}

\label{tab:unshuffled_results}
\end{table*}

\bigskip


\begin{table*}[hbtp]
\centering
\resizebox{0.55\textwidth}{!}{%
\begin{tabular}{lccc}
\toprule
\textbf{Base model + Method}              & \textbf{Unshuffled}$\downarrow$ & \textbf{Shuffled}$\downarrow$ & $\lvert\Delta\rvert\downarrow$ \\ 
\midrule
\textbf{DeepSeek V3} & & & \\
\quad Options (Original)                 & 66.70 & 67.02 & 0.32 \\
\quad Options (Aug by LLM Directly)      & 59.22 & 59.02 & 0.20 \\
\rowcolor{blue!6}
\quad Options (\ours{})                  & 57.06 & 56.92 & 0.14 \\[0.5ex]
\textbf{Qwen2.5-7B-Ins} & & & \\
\quad Options (Original)                 & 47.11 & 47.75 & 0.64 \\
\quad Options (Aug by LLM Directly)      & 42.16 & 41.34 & 0.82 \\
\rowcolor{blue!6}
\quad Options (\ours{})                  & 37.00 & 36.43 & 0.57 \\
\midrule
\bottomrule
\end{tabular}}
\caption{
Aggregate accuracy (\%, averaged across six datasets) before and after randomizing the answer-choice order.  
$\lvert\Delta\rvert$ is the \emph{absolute} difference between the shuffled and unshuffled scores.  
All $\lvert\Delta\rvert$ are below 1 percentage point which indicate no systematic benefit or harm from shuffling.  
Rows highlighted in blue correspond to our knowledge-guided distractor generation pipeline.  
\textbf{The main paper uses \textit{shuffled} KGGDG-augmented options as the default evaluation benchmark}.
}
\label{tab:shuffle_comparison}
\end{table*}

\end{document}